\documentclass[lettersize,journal]{IEEEtran}
\usepackage{amsmath,amsfonts}
\usepackage{algorithmic}
\usepackage{algorithm}
\usepackage{array}
\usepackage[caption=false,font=footnotesize,labelfont=sf,textfont=sf]{subfig}
\usepackage{textcomp}
\usepackage{stfloats}
\usepackage{url}
\usepackage{verbatim}
\usepackage{graphicx}
\usepackage{booktabs}
\usepackage{cite}
\usepackage{makecell}
\usepackage{multirow}
\begin{document}

\title{DFYP: A Dynamic Fusion Framework with Spectral Channel Attention and Adaptive Operator learning for Crop Yield Prediction}

\author{Juli Zhang,  Zeyu Yan, Jing Zhang, Qiguang Miao, Quan Wang
\thanks{This work was supported in part by the National Natural Science Foundations of China under grant No. 62272364, the provincial Key Research and Development Program of Shaanxi under grant No. 2024GH-ZDXM-47. (\textit{Corresponding authors: Juli Zhang, Qiguang Miao.})

Juli Zhang is with the School of Computer Science and Technology, Xidian University, Xi'an, 710065, China. She is also with the School of Computing, Australian National University, Canberra, ACT 2601, Australia. (email: zhangjuli@xidian.edu.cn, juli.zhang@anu.edu.au).

Zeyu Yan, Qiguang Miao and Quan Wang are with the School of Computer Science and Technology, Xidian University, Xi'an, 710065, China (email: 22009200335@stu.xidian.edu.cn, qgmiao@xidian.edu.cn, qwang@xidian.edu.cn).

Jing Zhang is with the School of Computing, Australian National University, Canberra, ACT 2601, Australia (e-mail: zjnwpu@gmail.com).

}
}
\markboth{Journal of \LaTeX\ Class Files,~Vol.~14, No.~8, August~2021}%
{Shell \MakeLowercase{\textit{et al.}}: A Sample Article Using IEEEtran.cls for IEEE Journals}

\IEEEpubid{0000--0000/00\$00.00~\copyright~2021 IEEE}

\maketitle

\begin{abstract}
Accurate remote sensing-based crop yield prediction remains a fundamental challenging task due to complex spatial patterns, heterogeneous spectral characteristics, and dynamic agricultural conditions. Existing methods often suffer from limited spatial modeling capacity, weak generalization across crop types and years. To address these challenges, we propose DFYP, a novel Dynamic Fusion framework for crop Yield Prediction, which combines spectral channel attention, edge-adaptive spatial modeling and a learnable fusion mechanism to improve robustness across diverse agricultural scenarios. Specifically, DFYP introduces three key components: (1) a \textit{Resolution-aware Channel Attention (RCA)} module that enhances spectral representation by adaptively reweighting input channels based on resolution-specific characteristics; (2) an \textit{Adaptive Operator Learning Network (AOL-Net)} that dynamically selects operators for convolutional kernels to improve edge-sensitive spatial feature extraction under varying crop and temporal conditions; and (3) a \textit{dual-branch architecture} with a learnable fusion mechanism, which jointly models local spatial details and global contextual information to support cross-resolution and cross-crop generalization. Extensive experiments on multi-year datasets MODIS and multi-crop dataset Sentinel-2 demonstrate that DFYP consistently outperforms current state-of-the-art baselines in RMSE, MAE, and R² across different spatial resolutions, crop types, and time periods, showcasing its effectiveness and robustness for real-world agricultural monitoring.
\end{abstract}

\section{Introduction}
\label{sec:intro}

Accurate crop yield prediction plays a critical role in ensuring global food security and informing agricultural decision-making. However, this task remains highly challenging due to the complex interplay of meteorological conditions, soil properties, crop types, and cultivation practices~\cite{aghighi2018machine, elavarasan2018forecasting}. Furthermore, yield prediction is a nonlinear and dynamic process~\cite{qiao2023kstage}—integrating multi-source signals that evolve over time and vary across spatial regions—making robust prediction even more difficult.

Deep learning has emerged as a powerful tool for crop yield prediction~\cite{liu2019crop,jabed2024crop,shahhosseini2021corn}, enabling automated feature extraction from remote sensing imagery and offering substantial improvements over traditional statistical and empirical models. However, despite its promise, deep learning-based yield prediction still faces critical challenges in real-world agricultural applications. \textit{1) Neglect of spatial dependencies}: Many models treat spatial locations independently~\cite{helber2023crop,you2017deep,khaki2020cnn}, ignoring important correlations across regions, which reduces accuracy in large-scale predictions. \textit{2) Limitations of single architectures}: CNN-based methods~\cite{shahhosseini2021corn, khaki2020cnn, zhang2016deep} are effective for extracting local features but rely on fixed kernels, limiting adaptability under environmental changes~\cite{aleissaee2023transformers}. ViTs~\cite{dosovitskiy2020image}, on the other hand, excel at modeling global dependencies, yet are prone to overfitting with high-resolution imagery and often miss fine-grained spatial details~\cite{ma2019deep, graham2021levit}. \textit{3) Lack of generalization across crops and datasets}: Most existing studies are restricted to single-crop or single-region scenarios~\cite{khaki2019crop,schwalbert2020satellite,lu2024deep}, limiting their utility in diverse agricultural environments. \textit{4) Inflexible edge feature modeling}: Edge structures in remote sensing imagery are critical for distinguishing crop boundaries~\cite{zhou2022edge}. However, conventional CNNs do not adapt well to the temporal or crop-specific variability of edge characteristics~\cite{le2021revisiting}. Our empirical study across eight classical edge operators reveals that their performances fluctuate significantly with crop type (Sentinel-2) and acquisition year (MODIS), which indicates that fixed operators are suboptimal for robust crop yield prediction.

Real-world crop yield prediction often demands robustness across heterogeneous resolutions, temporal variations, and multiple crop types. Satellite imagery from different sensors (e.g., MODIS or Sentinel-2) exhibits significant diversity in spatial, spectral, and temporal properties. Additionally, seasonal changes, crop rotations, and environmental dynamics introduce concept drift~\cite{lu2018learning,gama2014survey}, challenging model generalization over time. Existing approaches~\cite{zhang2016deep,kamilaris2018deep,oikonomidis2023deep,khaki2019crop,jhajharia2023crop,shahhosseini2021corn,khaki2020cnn} are insufficient to handle such diverse agricultural conditions. These challenges highlight the critical need for robust, flexible, and generalizable yield prediction frameworks that can effectively adapt to heterogeneous data resolutions, spectral variability, and complex spatiotemporal dynamics in real-world agricultural environments.

To overcome these challenges, we propose DFYP, a Dynamic Fusion framework for Yield Prediction that combines Resolution-aware Channel Attention and Adaptive Operator Learning. The design introduces three key innovations: 

\IEEEpubidadjcol
\begin{itemize}
    \item A Dual-branch dynamic fusion architecture is proposed to optimally balance local and global feature learning through a learnable fusion mechanism. This enables the model to capture both short-range and long-range spatial dependencies.
   \item Adaptive Operator Learning Network(AOL-Net) is designed for concept drift, which dynamically selects the most effective edge operator (e.g., Sobel, Scharr, or a learned operator) based on the input’s temporal and semantic characteristics, thus improving spatial robustness across years and crop types.
   \item A Resolution-aware Channel Attention (RCA) module is introduced to adaptively reweight spectral bands based on image resolution. This  design enhances the model’s ability to generalize across heterogeneous sensors and crop types by selectively emphasizing the most informative spectral features under both high- and low-resolution conditions.
\end{itemize}

The remainder of this paper is constructed as follows. Section II reviews the related work of crop yield prediction and hybrid models, and the need for dynamic fusion. Section III describes the details of the proposed method. The experimental results and discussions are presented in Section IV. We conclude this work in Section V.
\section{Related Work}
\label{sec:related work}
\subsection{Deep Learning for Crop Yied Prediction}
The repid development of deep learning in computer vision task~\cite{lecun2015deep,redmon2016you} has catalyzed its widespread application in agricultural domains, particularly for crop yield prediction~\cite{zhang2016deep,kamilaris2018deep,oikonomidis2023deep,khaki2019crop,jhajharia2023crop,shahhosseini2021corn,khaki2020cnn}. Existing approaches generally fall into two categories: remote sensing-based approaches ~\cite{khaki2021simultaneous,khaki2019crop,lees2022deep,garnot2021panoptic}, which leverages satellite imagery, unmanned aerial vehicle (UAV) data, or vegetation indices; and meteorological-based approaches~\cite{khaki2020cnn,fan2022gnn,akhavizadegan2021time,mourtzinis2021advancing,turchetta2022learning}, which model the impact of climatic variables using deep  neural networks (DNNs).

Among DNN-based models, convolutional neural networks (CNNs) have shown strong capabilities in automatically extracting spatial features from satellite images~\cite{nevavuori2019crop,chen2014deep,tyagi2024enhancing,bhatt2017comparison}. However, CNNs are inherently limited by their local receptive fields, which restrict their ability to model long-range spatial dependencies~\cite{aleissaee2023transformers,yunusa2024exploring}, reducing effectiveness in large-scale agricultural landscapes. To address this,  ViTs~\cite{dosovitskiy2020image,zamir2022restormer,vaswani2017attention,zhang2019self,khan2022transformers} have been explored for their strength in capturing global spatial relationships. Recent studies have demonstrated the potential of ViTs in land cover classification~\cite{yao2023extended} and crop yield prediction~\cite{lin2023mmst,pacal2024enhancing}. Nevertheless, ViTs often require substantial amounts of training data and may struggle to preserve fine-grained spatial details~\cite{peng2021conformer}, especially when dealing with high-resolution imagery. This limits their standalone utility in complex agricultural environments.

Recent models have attempted to address these limitations. For example, DeepField~\cite{Gavahi2021} uses a CNN backbone to predict yields from temporal image sequences but lacks adaptive fusion. MMST-ViT~\cite{lin2023mmst} incorporates multi-modal transformer modules but applies them statically without resolution-aware adjustments. These methods all rely solely on CNN or ViT for modeling, and still have certain limitations in their ability to model both the spatial and temporal characteristics of the data. These limitations highlight the need for hybrid frameworks that can balance local spatial sensitivity and global contextual modeling for robust yield prediction.

\subsection{Hybrid models and The Need for Dynamic Fusion}
Building on the complementary nature of CNNs and ViTs, studies have proposed hybrid architectures~\cite{wu2021cvt,yuan2021incorporating,hassani2021escaping,li2021localvit,graham2021levit,zhang2021aggregating,carion2020end,peng2021conformer} that combine CNNs’ ability to extract local spatial features with the global context modeling capabilities of Transformers. According to the survey~\cite{yunusa2024exploring}, these hybrid models can be categorized into several types: 1) parallel integration: Conformer(~\cite{peng2021conformer}, Mobile-Former~\cite{chen2022mobile}, TCCNet~\cite{li2022tccnet}, 2) serial combination: CETNet~\cite{wang2024cross}, CoATNet~\cite{dai2021coatnet}, CMT~\cite{guo2022cmt}, 3) hierarchical architectural integration: ViTAE~\cite{xu2021vitae}, CVT~\cite{wu2021cvt}, DiNAT~\cite{hassani2022dilated}, 4) early stage fusion: DETR~\cite{carion2020end}, MASK Former~\cite{cheng2021per}, LeViT~\cite{graham2021levit}, 5) late-stage fusion: Efficient Former~\cite{li2022efficientformer}, Swin2SR~\cite{conde2022swin2sr}, ViTMatte~\cite{yao2024vitmatte}, and 6) attention-based fusion: HBCT~\cite{batalik2021long}, ECA-NET~\cite{wang2020eca}, DANet~\cite{fu2019dual}, scSE~\cite{roy2018concurrent}, CBAM~\cite{woo2018cbam}, BoTNet~\cite{srinivas2021bottleneck}. While these hybrid approaches combine CNN and ViT features across various stages and layers, they generally rely on static integration strategies such as feature concatenation, averaging, or fixed attention mechanisms, which do not dynamically adjust the contributions of CNN and ViT features based on data-specific characteristics. These static approaches, particularly in straightforward concatenation, treat each feature equally and fail to account for the varying relevance of local versus global features across different contexts, reducing their flexibilities. Consequently, the lack of adaptive feature weighting diminishes the effectiveness of hybrid models, especially in complex remote sensing applications where feature importance fluctuates depending on spatial and spectral conditions.

Our approach introduces a loss-guided fusion mechanism that dynamically learns to weight the outputs of the AOL-Net and ViT branches based on training data. This adaptive fusion enables better alignment with complex spatiotemporal variations in remote sensing imagery.

\subsection{Traditional Model Combinations for Yield Prediction}
Beyond hybrid deep models,  a variety of classical model combinations have been employed in crop yield prediction. These approaches typically integrate spatial and temporal modeling in a modular fashion. In CNN-RNN~\cite{khaki2020cnn} and CNN-LSTM~\cite{Nevavuori2020}, CNNs are used to extract spatial features from remote sensing imagery, which are then passed to RNN or LSTM layers for modeling temporal dynamics across growing seasons. GNN-RNN~\cite{fan2022gnn} combines Graph Neural Networks for spatial representation and recurrent layers for modeling temporal dynamics. 3DCNN-convLSTM~\cite{Nejad2023} utilizes 3D-CNNs to simultaneously extract spatial and short-term temporal features from image sequences, which are then refined through ConvLSTM modules for enhanced sequential prediction. 4) ConvLSTM + ViT~\cite{mirhoseininejadConvLSTMViTDeep2024} combines ConvLSTM with ViT to jointly model sequential dynamics and global spatial context, though such designs often rely on static feature integration and lack adaptive fusion mechanisms.  GP-based models (Gaussian Process)~\cite{you2017deep} incorporate a GP regression layer after feature extraction for uncertainty-aware prediction or refined output modeling.

While effective in specific tasks, these combinations generally require separate training for spatial and temporal modules, limiting end-to-end optimization and complicating backpropagation. Additionally, they often rely on handcrafted feature engineering or static architectures that fail to generalize across datasets with diverse spatial-spectral characteristics. Furthermore, they tend to exhibit significant performance fluctuations across different crop types and temporal domains, limiting their robustness in real-world yield prediction scenarios.

In contrast, DFYP offers a robust end-to-end framework that combines adaptive spectral reweighting (RCA), dynamic edge-aware CNNs (AOL-Net), and Transformer-based global modeling with flexible. Furthermore, the dynamic loss-driven fusion leads to improved generalizability across crop types, time periods, and sensor resolutions.

\section{Methodologies}
\begin{figure*}
  \centering
    \includegraphics[width=0.8\linewidth]{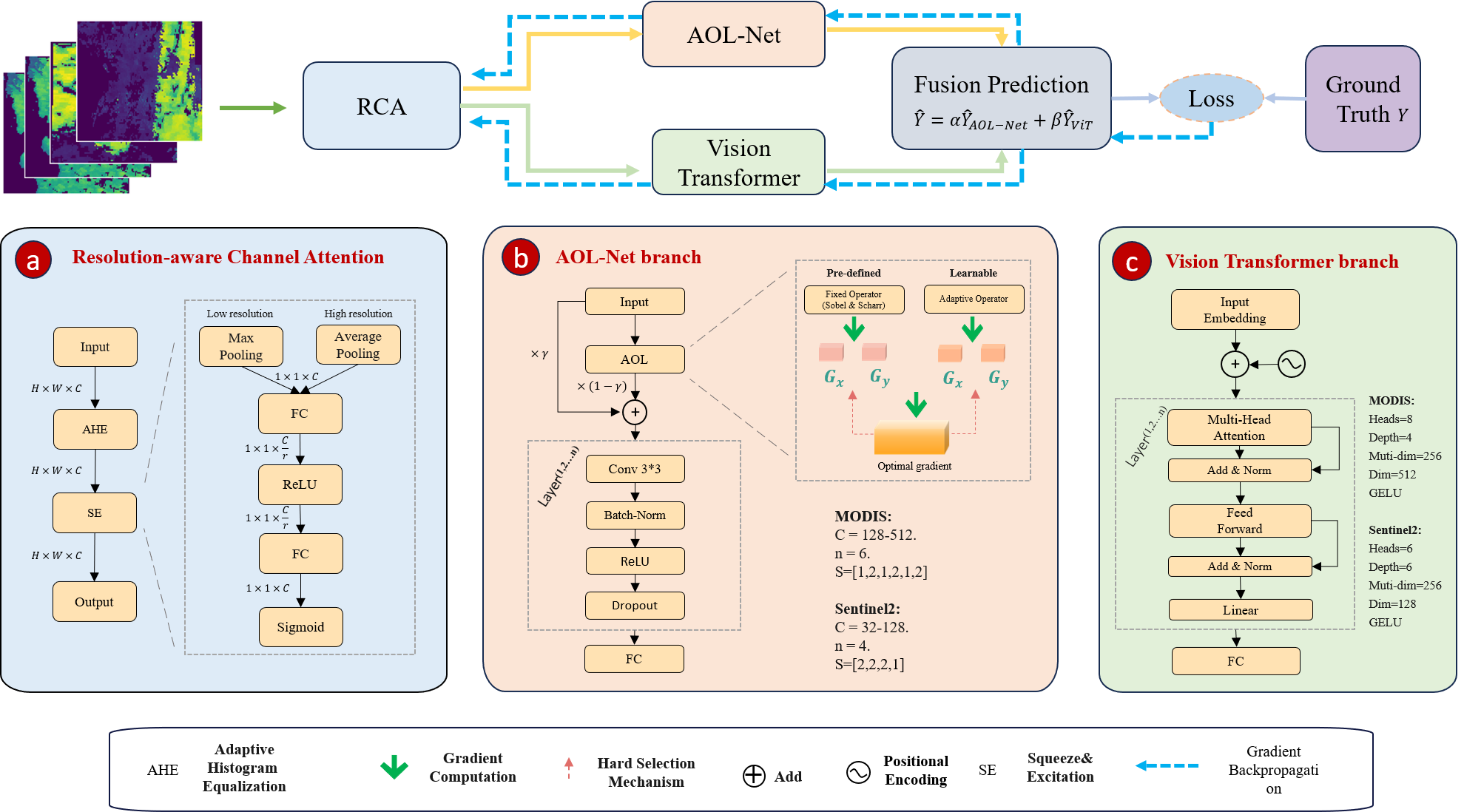}
    \caption{Overview of the proposed DFYP framework for remote sensing-based crop yield prediction. The input consists of multi-spectral remote sensing imagery, which is first processed by the \textbf{a) Resolution-aware Channel Attention (RCA)} module to adaptively enhance spectral channel importance based on spatial resolution. The refined features are then fed into two parallel branches: the \textbf{b) AOL-Net branch}, which utilizes an Adaptive Operator Library for edge-aware convolution, and the ViT branch, which models long-range dependencies via a \textbf{c) Vision Transformer}.Each branch outputs predictions through a fully connected (FC) layer. These predictions are fused via a learnable weighted sum to form the final prediction. The total loss is computed between the fused prediction and the ground truth, with gradients backpropagated to both branches for end-to-end optimization.}
    \label{fig:Overview of DFYP.}
\end{figure*}
In this work, we aim to develop a unified and robustcrop yield prediction framework, which can handle remote sensing imagery with different resolutions. This framework should achieve three goals: (1) learning resolution-aware spectral representations that adapt to differing input characteristics across low- and high-resolution imagery; (2) enhancing spatial and temporal feature extraction from dynamic agricultural conditions, and (3) fusing local and global representations via a learnable mechanism. Our ultimate objective is to build a model that maintains predictive performance across diverse crops, years, and sensor types, while remaining end-to-end trainable and interpretable.

\subsection{Problem Definition}
Given a dataset \(\mathcal{D} = \{(\mathbf{X}_i, Y_i)\}_{i=1}^{N}\) consisting of remote sensing images and their corresponding crop yield labels, our goal is to learn a predictive function \( f_{\theta} \) that maps input features \( \mathbf{X} \) to crop yield predictions \( \hat{Y} \):
\begin{equation}
\label{eq_main_obj}
    \hat{Y} = f_{\theta}(\mathbf{X}) + \epsilon,
\end{equation}
where \( \mathbf{X} \in \mathbb{R}^{C \times H \times W} \) is an image with \textbf{\( C \)} spectral channels and spatial dimensions \( H \times W \), \( \hat{Y} \) represents the predicted crop yield, and \( \epsilon \) denotes the residual error. $f_{\theta} (\mathbf{X})$ is our prediction function. \( \theta \) are optimized to minimize the prediction error. 

\subsection{Overview of the Proposed Approach}
To achieve the aforementioned goals, we proposed DYFP. As illustrated in Figure \ref{fig:Overview of DFYP.}, DFYP comprises three key modules: a spectral attention encoder, a dual-branch representation extractor, and a dynamic fusion module.

The first component, Resolution-Aware Channel Attention module, aims to adaptively reweight multispectral channels according to their informativeness across different sensors and crop types. By encoding spectral saliency in a resolution-aware manner, this module improves the robustness of feature representations under heterogeneous input conditions.

Next, DFYP employs a dual-branch learning architecture, consisting of a Adaptive Operator Learning (AOL) branch and a parallel Vision Transformer (ViT) branch. The AOL branch dynamically selects the most appropriate edge enhancement operator for each input. This design improves the model's robustness to concept drift and helps preserve critical spatial structures across different crop types and seasons. In parallel, the ViT branch captures long-range spatial dependencies, which are often missed by purely convolutional models.

Finally, the outputs of both branches are subsequently fused through a learnable mechanism for final prediction, which adaptively balances local and global information to produce the final yield prediction. This design ensures that DFYP remains flexible across regions, crops, and years, and maintains stability even under extrem weather conditions such as drought or sensor noise.

Detailed descriptions of each module are provided in the following subsections.

\subsection{Resolution-aware Channel Attention}

To enhance the model’s ability to selectively emphasize informative spectral features across heterogeneous remote sensing data sources, we introduce the Resolution-aware Channel Attention (RCA) module. This module is inspired by the classic Squeeze-and-Excitation (SE) network~\cite{hu2018squeeze}, which models inter-channel dependencies via global pooling and gating. However, unlike SE, which applies a uniform strategy regardless of image modality, RCA explicitly incorporates spatial resolution awareness into the attention computation, making it particularly suitable for different multispectral imagery such as MODIS (500m) and Sentinel-2 (10m).

To handle spectral heterogeneity across resolutions, we design the channel attention mechanism with a resolution-aware pooling strategy tailored for multi-resolution remote sensing imagery. RCA consists of three stages: \textit{Squeeze}, \textit{Excitation}, and \textit{Reweighting}.

\noindent\textbf{(a) Squeeze Operation (Resolution-aware Pooling).}  
To aggregate global information, we replace the fixed pooling in standard SE blocks with an adaptive pooling mechanism:
\begin{equation}
  z_c = 
  \begin{cases} 
  \max\limits_{u,v} X_c(u, v), & \text{if low resolution} \\
  \frac{1}{H \times W} \sum\limits_{u,v} X_{c}(u, v), & \text{if high resolution}
  \end{cases}
\end{equation}
where \( z_c \) is the aggregated channel descriptor, dynamically computed using max pooling for low-resolution images to emphasize key structures and average pooling for high-resolution images to preserve information integrity.

\noindent\textbf{(b) Excitation Operation (Channel-wise Importance Estimation).}  
Feature \( z_c \) is passed through two fully connected layers with ReLU and Sigmoid activations to generate channel-wise importance scores: $s_c = \sigma(W_2 \delta(W_1 z_c))$,
where \( W_1 \in \mathbb{R}^{\frac{C}{r} \times C} \) and \( W_2 \in \mathbb{R}^{C \times \frac{C}{r}} \), \( r \) is a reduction ratio, \( \delta(\cdot) \) and \( \sigma(\cdot) \) denote  ReLU and  sigmoid functions, respectively.

\noindent\textbf{(c) Channel Reweighting.}  
Finally, the learned weights are applied to the original feature map: $\tilde{X}_c = s_c \cdot X_c$.
which adaptively enhances important spectral bands while suppressing irrelevant ones. 

By integrating resolution-awareness into channel attention, RCA improves cross-resolution generalization and enhances the robustness of feature representations for yield prediction.

\subsection{Adaptive Operator Learning Network}
CNNs use convolutional kernels with fixed receptive patterns, which inherently lack the flexibility to capture diverse and evolving edge structures in remote sensing imagery. These edge patterns vary significantly across different geographic regions, crop types, and acquisition periods due to changes in vegetation phenology, atmospheric conditions, and spectral characteristics. Although handcrafted edge operators (e.g., Sobel, Scharr) can emphasize gradients, their performance is highly sensitive to crop types, seasonal changes, and spectral inconsistencies. Our empirical experiments across multi-year and multi-crop datasets demonstrate that no single operator consistently performs well across all conditions. This motivates the need for a dynamic mechanism that adapts edge enhancement to the input context. 

To address these challenges, we introduce an Adaptive Operator Learning Network (AOL-Net), a dedicated module for robust edge-aware feature enhancement. Unlike conventional CNNs that rely on static convolutional filters, AOL-Net dynamically adjusts the edge encoding mechanism through a learnable operator fusion strategy, inspired by the residual learning principle. AOL-Net is composed of the following components:

\noindent\textbf{(a) Operator Pool.} A predefined library  $\mathcal{K} = \{K_{\text{Sobel}}, K_{\text{Scharr}}, K_{\text{Learnable}}\}$, where $K_{\text{Sobel}}, K_{\text{Scharr}}$ are fixed classical operators, $K_{\text{Learnable}}$ is a trainable kernel initialized via a convex combination:
\begin{equation}
\begin{split}
 K_{\text{learnable}} = \lambda K_{\text{Sobel}} + (1-\lambda) K_{\text{Scharr}}.\quad 0 \leq \lambda \leq 1  
\end{split}
\end{equation}
\noindent\textbf{(b) Residual-inspired Fusion.}
During training, \( \lambda \)  is optimized via backpropagation to automatically interpolate between fixed operators or revert to them when beneficial. When  \( \lambda \) converges to 0 or 1, AOL effectively reduces to a traditional operator Sobel or Scharr, preseving stability. This mechanism mimics residual connections by granting the network the ability to retain, refine, or bypass edge operations.

\noindent\textbf{(c) Operator Selection Gate.}
The final operator applied at a given time step \( t \) is determined by a selection function:
\begin{equation}
    K_t = \arg\max_{K \in \mathcal{K}} \mathcal{S}(K, t)
\end{equation}
where \( \mathcal{S}(K, t) \) represents the historical performance score of each operator, computed based on validation results: $\mathcal{S}(K, t) = \frac{1}{T} \sum_{\tau=1}^{T} \text{Accuracy}(K, \tau)$. 
 \( K_t \) is then applied to the input image to extract edge features: $G_t = K_t \ast X$,
where \( G_t \) is the resulting edge-enhanced feature map.

Instead of soft selection, we employ hard selection:
\begin{equation}
\begin{split}
      \lambda_{\text{Sobel}}, \lambda_{\text{Scharr}}, \lambda_{\text{Learnable}} \in \{0,1\},\\ \quad
    \lambda_{\text{Sobel}} + \lambda_{\text{Scharr}} + \lambda_{\text{Learnable}} = 1  
\end{split}
\end{equation}
where exactly one coefficient is active per time step. The final operator used for edge detection is:
\begin{equation}
    K_t = \lambda_{\text{Sobel}} K_{\text{Sobel}} + \lambda_{\text{Scharr}} K_{\text{Scharr}} + \lambda_{\text{Learnable}} K_{\text{Learnable}}
\end{equation}
\noindent\textbf{(d) Embedding into Backbone.}  $K_t$ is used to modulate the input feature map before convolution, enabling the backbone to emphasize edge-aware structures during feature extraction.
\begin{equation}
X' = \gamma \cdot X + (1 - \gamma) \cdot G_t
\end{equation}
\begin{equation}
Y = W * X’ + b
\end{equation}
where \( W \) represents learnable convolution weights, \( \gamma \) is a learnable parameter that adaptively scales the influence of the edge operator, and \( b \) is a bias term. It ensures that the CNN dynamically incorporates adaptive edge information into its feature extraction process.


\subsection{Global Spatial Modeling with Vision Transformer}
While RCA and AOL-Net modules focus on spectral selection and edge enhancement respectively, the Vision Transformer (ViT) branch is responsible for modeling long-range spatial dependencies critical to yield prediction. Each RCA-enhanced image is partitioned into fixed-size patches and embedded into a latent space via a linear projection layer. We adopt positional encodings to preserve spatial ordering and feed the embedded sequence into multiple Transformer layers to capture non-local context.

\noindent\textbf{Patch Embedding and Input Representation.}  
The input image \( X \) is divided into non-overlapping patches and projected into an embedding space:
\begin{equation}
X_{\text{proj}} = XW_{\text{emb}} + b_{\text{emb}},
\end{equation}
where \( W_{\text{emb}} \) and \( b_{\text{emb}} \) denote the embedding weight matrix and bias. Positional encodings \( P \) are then added to retain spatial relationships.

\noindent\textbf{Transformer Encoding.}  
The position-enhanced embeddings are passed through multiple Transformer layers, where each layer models interactions via Multi-Head Attention (MHA):
\begin{equation}
\text{Attention}(Q, K, V) = \text{softmax}\left( \frac{QK^T}{\sqrt{d_k}} \right) V.
\end{equation}
The output is processed using residual connections and a feed-forward network (FFN):
\begin{equation}
X_{\text{out}} = \text{FFN}(X + \text{MHA}(X)).
\end{equation}

\noindent\textbf{Final Prediction.}  
After Transformer encoding, the final feature representation is passed through an MLP head to generate yield predictions. This global feature extraction allows the model to learn high-level contextual relationships that are crucial for yield estimation.
\subsection{Loss-Guided Fusion Mechanism}
 After generating crop yield predictions independently from the AOL-Net and ViT branches, we introduce a learnable fusion mechanism to combine their outputs.

\noindent\textbf{Fusion Strategy.}  According to Equation 1, our objective is to minimize the error between the predicted yield $\hat{Y}$ and the actual yield $Y$. Since our AOL-Net + ViT framework adopts a dynamic fusion strategy, we define $f_{\theta} (\mathbf{X})$ as:

\begin{equation}
f_{\theta} (\mathbf{X}) = \alpha f_{\theta_1} (\mathbf{X}) + \beta f_{\theta_2} (\mathbf{X})
\end{equation}
where $f_{\theta_1} (\cdot)$ and $f_{\theta_2} (\cdot)$ are the outputs of the AOL-Net and ViT branches, respectively. The fusion weights $\alpha, \beta  \in(0,1)$ are learnable parameters that are jointly optimized with the rest of the network via backpropagation.

\noindent\textbf{Loss Function.} 
We adopt the Mean Squared Error (MSE) as the objective:
\begin{equation}
L_{\text{total}} = \frac{1}{N} \sum_{i=1}^{N} (\hat{Y}_i - Y_i)^2
\end{equation}

Substituting \( \hat{Y}_i = \alpha f_{\theta_1} (\mathbf{X}_i) + \beta f_{\theta_2} (\mathbf{X}_i) \), we obtain:
\begin{equation}
\begin{split}
    L_{\text{total}} = \frac{1}{N} \sum_{i=1}^{N} (\alpha f_{\theta_1} (\mathbf{X}_i) + \beta f_{\theta_2} (\mathbf{X}_i) - Y_i)^2
\end{split}
\end{equation}
Unlike normalized fusion schemes that enforce $\alpha +\beta=1$, we relax this constraint to allow both $\alpha$ and $\beta$ to be learned independently with $(0,1)$. This design offers greater flexibility by enabling the model to adaptively prioritize either branch or maintain a balance between them, depending on the data distribution. Empirical results have shown that this approach improves robustness and ensures more stable convergence during training.

All modules in DFYP are trained jointly in an end-to-end fashion. During training, the final prediction is obtained by fusing outputs from both branches, and the overall loss is backpropagated through the entire architecture, allowing each component to adapt to the input distribution and optimize its parameters accordingly.

\section{Experiment and Discussion}
\label{experiment}
\subsection{Experimental Settings}

\noindent\textbf{Datasets.}  
We conduct experiments on two remote sensing datasets: Moderate Resolution Imaging Spectroradiometer (MODIS) and Sentinel-2 imagery. MODIS is a key instrument of the Earth Observing System (EOS) onboard NASA’s Terra satellite, launched in 1999, and Aqua satellite, launched in 2002. In this study, we utilize MODIS-derived data, which includes three components: surface reflectance, surface temperature, and soil cover type, which were accessed via Google Earther Engin. Sentinel-2 images we utilized in this work were from the dataset described in the literature~\cite{lin2024open}. In this work, we focus on soybean yield prediction on MODIS, while we cover four major crops—corn, cotton, soybean, and winter wheat on Sentinel-2. To facilitate a more comprehensive understanding of our data, we provide detailed descriptions and preprocessing procedures in the appendix.

\noindent\textbf{Experimental Setup.}  
To predict yield for a given year \( t \), the models are trained using data from \( t_{\text{start}} \) to \( t-1 \), with 90\% of the multi-spectral remote sensing data used for training and 10\% for validation. For MODIS-based experiments, models are trained using data from $t_{\text{start}}$ = 2003 to t - 1, and evaluated on the years t = \{2009\text{–}2015\}. For Sentinel-2 imagery, we use data from $t_{\text{start}}$ = 2017 and evaluate on t = 2022. All Predictions rely solely on pre-season and in-season observations, without access to post-harvest data. Evaluation is conducted at the county level using official yield statistics from the United States Department of Agriculture (USDA).  

\begin{table*}[t]
\centering
\setlength{\tabcolsep}{4pt} 
\renewcommand{\arraystretch}{1.2} 

\caption{Hyperparameter settings for our experiments}
\label{tab:hyperparams}
\begin{tabular}{>{\raggedright\arraybackslash}p{1.8cm} ccccccccccc}
\toprule
\textbf{Dataset} & \textbf{Opt.} & \textbf{LR} & \textbf{Steps} & \textbf{Batch} & \textbf{CNN Channels} & \textbf{Strides} & \textbf{Depth} & \textbf{Heads} & \textbf{Dim} & \textbf{MLP} & \textbf{Image \& Patch} \\
\midrule
MODIS & Adam & 1e-4 & 25K & 64 & [in, 128, 256, 256, 512, 512, 512] & [1,2,1,2,1,2] & 4 & 8 & 256 & 512 & \makecell{$32 \times 32$ \\ $4 \times 4$} \\
\midrule
Sentinel-2 & Adam & 1e-4 & 25K & 16 & [in, 32, 64, 128, 128] & [2,2,2,1] & 6 & 6 & 128 & 256 & \makecell{$256 \times 256$ \\ $16 \times 16$} \\
\bottomrule
\end{tabular}

\vspace{0.7em}
\begin{minipage}{0.96\textwidth}
\footnotesize
\textit{Notes:} Opt.: optimizer, LR: learning rate, Dim: embedding dimension, MLP: MLP dimension. ReLU activation is used in CNN layers while GELU is used in Transformer blocks. Early stopping patience of 10 is applied to both datasets. “in” represents input channel dimension.
\end{minipage}
\end{table*}

\noindent\textbf{Implementation Details.}
DFYP is implemented in PyTorch, with dataset-specific configurations tailored to the distinct properties of MODIS and Sentinel-2. Both experiments adopt a dual-branch architecture comprising independent CNN and ViT modules. MODIS provides coarse-resolution, temporally frequent imagery, whereas Sentinel-2 offers high-resolution, multi-spectral data. These differences motivate distinct configurations:

\vspace{1mm}
\textit{1) AOL-Net Module:}
The CNN component is adjusted according to the resolution of each dataset:
\begin{itemize}
\item For \textbf{MODIS}, a deeper CNN is employed to compensate for limited detail by learning higher-level abstractions. Each layer consists of a 3×3 convolution, ReLU activation, batch normalization, and dropout for regularization.
\item For \textbf{Sentinel-2}, a shallower CNN suffices due to the input's inherent richness. This design avoids potential overfitting and maintains computational efficiency.
\end{itemize}

\vspace{1mm}
\textit{2) ViT Module.}
The Vision Transformer branch comprises a linear projection, positional embeddings, multiple MHSA layers, FFNs, and layer normalization. Its configuration is similarly dataset-aware: For MODIS, a lighter encoder is used to model global dependencies without overfitting to coarse details.
For Sentinel-2, it utilizes a deeper encoder to better capture complex spatial relationships inherent in high-resolution imagery.

All models are trained using the Adam optimizer with early stopping based on validation loss. Hyperparameters for each variant are summarized in Table~\ref{tab:hyperparams}.

\vspace{1mm}
\noindent
This dataset-aware design ensures that DFYP effectively captures both local textures and global structures under diverse remote sensing conditions, maintaining a balance between representational capacity and generalization.

\begin{figure*}[t]
  \centering
    \includegraphics[width=1\linewidth]{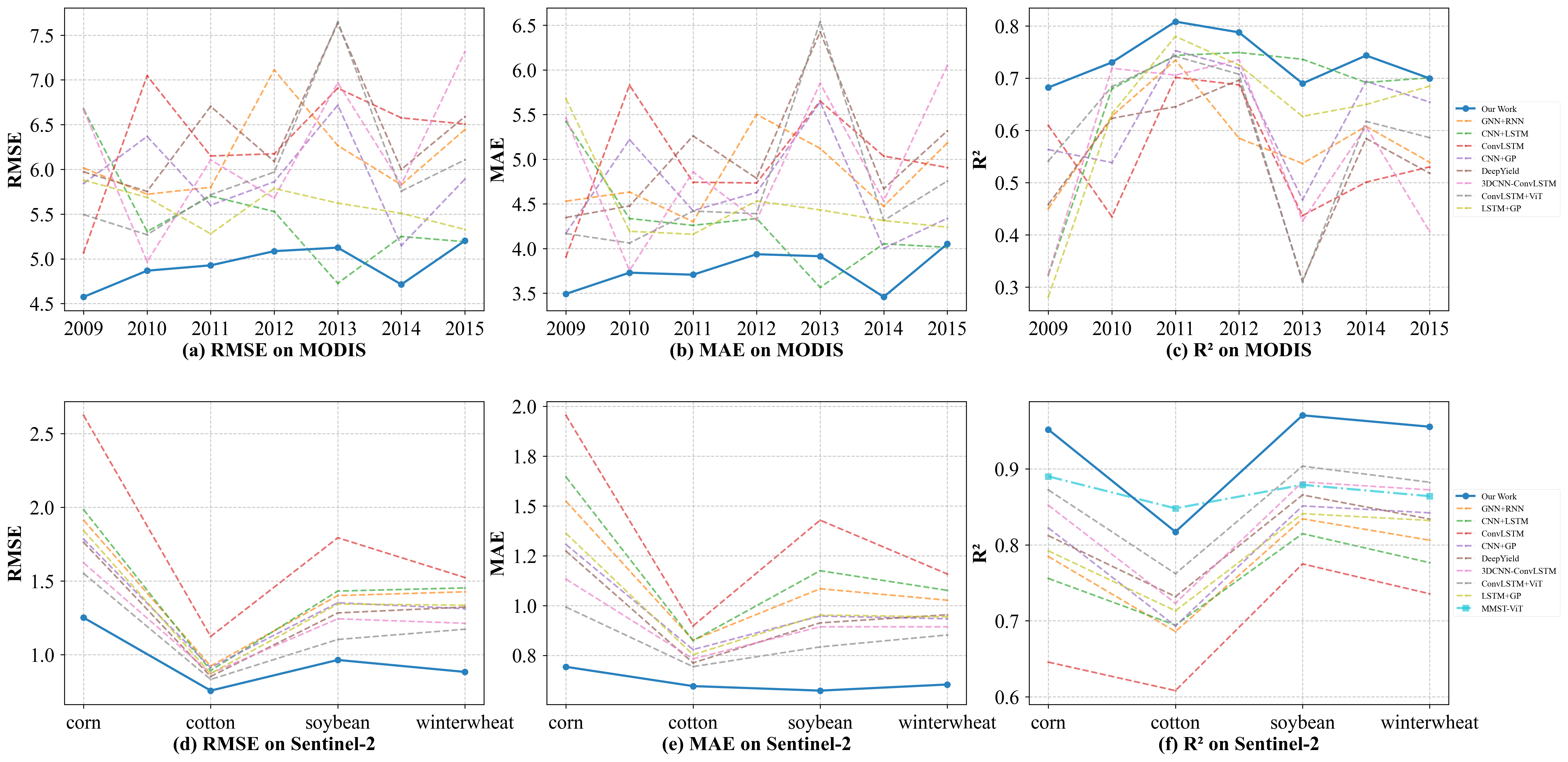}
    \label{fig:MAE and RMSE of Corn yield on the Sentinel-2 Data.}
  \caption{Comparision with state-of-the-art methods. Fig. (a, b, c) show the RMSE, MAE and $R^2$ performance on MODIS dataset. Fig. (d, e, f) show the performance on Sentinel 2 dataset.}
  \label{fig: comparision with sota methods.}
\end{figure*}

\noindent\textbf{Compared Approaches.} 
To prove the superiority of the proposed DFYP, we compare it with  the following state-of-the-art (SOTA) approaches:

(1) GNN+RNN~\cite{fan2022gnn} captures both spatial and temporal dependencies for crop yield prediction. It encods yearly environmental features with CNNs, aggregates spatial context via GraphSAGE, and models inter-annual trends using LSTMs.

(2) CNN+LSTM~\cite{Nevavuori2020} sequentially applies a CNN for spatial feature extraction and an LSTM for temporal modeling using UAV-based RGB time series data.

(3) ConvLSTM~\cite{Guo2021} models spatiotemporal dependencies by replacing fully connected operations in LSTM with convolutions, enabling joint learning of spatial and temporal patterns from gridded time series data.

(4) CNN+GP~\cite{you2017deep} learns spatial features from histogram-transformed remote sensing data using a CNN, and incorporates a Gaussian Process layer to model spatio-temporal correlations and reduce prediction errors.

(5) Deepfield~\cite{Gavahi2021} integrates the ConvLSTM layers with the 3-Dimensional CNN (3DCNN) for more accurate and reliable spatiotemporal feature extraction.

(6) 3DCNN+ConvLSTM~\cite{Nejad2023} leverages 3D convolutions for spatio-spectral feature extraction and integrates attention-enhanced ConvLSTM to capture temporal dependencies in multi-temporal multispectral imagery.

(7) ConvLSTM+ViT~\cite{mirhoseininejadConvLSTMViTDeep2024} combines ConvLSTM for short-term spatiotemporal modeling with Vision Transformer to capture long-range spatial dependencies from imagery.

(8) LSTM+GP~\cite{you2017deep} uses LSTM to model temporal sequences of histogram-encoded satellite observations, while a Gaussian Process layer captures residual spatial and temporal dependencies for enhanced yield prediction.

(9) MMST-ViT ~\cite{lin2023mmst} combines satellite imagery and meteorological data through a Vision Transformer—including Multi-Modal, Spatial, and Temporal Transformers—to capture both short-term weather variation and long-term climate effects for accurate county-level crop yield prediction.

With the exception of MMST-ViT, which requires multimodal data, the remaining methods were rigorously replicated using the original papers' provided codes or methodologies. We utilize Fig. \ref{fig: comparision with sota methods.} to demonstrate the comparison results.

\noindent\textbf{Evaluation Metrics.} All methods are evaluated using three popular metrics: mean absolute error (MAE), root mean square error (RMSE), and coefficient of determination (R²). Importantly, model performance is considered better when the RMSE is lower and the R² is higher.
\subsection{Comparison with State-of-the-Art Methods}
We utilize Fig. \ref{fig: comparision with sota methods.} to demonstrate the comparision results. From Fig. \ref{fig: comparision with sota methods.}, we can observe that our approach consistently outperforms all baselines across RMSE and MAE on the Sentinel-2, and achieve the best $R^2$ except MMST-ViT on cotton, demonstrating its superior effectiveness in high-resolution satellite imagery. On the MODIS dataset, our method also achieves the best performance in six out of seven years (2009, 2010, 2011, 2012, 2014, 2015) across all metrics. The following sections provide detailed comparisons and further analysis of these results.


\noindent \textbf{Analysis on MODIS.} Notably, DFYP demonstrates superior performance and remarkable temporal stability across all three metrics, significantly outperforming other approaches in most years. The stable and consistently high performance of DFYP can be attributed to its hybrid design: ViT enables the modeling of long-range spatial dependencies, while AOL-CNN adaptively enhances edge features across time through operator selection. This dual-branch design mitigates the effects of concept drift, sensor noise, and spectral inconsistencies, all of which are prevalent in MODIS due to its coarse spatial resolution. In contrast, baseline models such as CNN+GP, DeepYield, and 3DCNN+ConvLSTM exhibit large fluctuations over time, with performance degrading sharply in 2013, highlighting limited robustness to interannual spectral and environmental variations. 

Among all models, CNN-LSTM slightly outperforms our method on all three metrics in 2013. This suggests that recurrent architectures, with their ability to exploit sequential continuity, may have been particularly effective in modeling year-to-year recovery patterns embedded in the temporal evolution of vegetation indices. This anomaly likely reflects the advantage of recurrent architectures in modeling localized temporal continuity—potentially aligning well with crop recovery patterns following the severe 2012 drought~\cite{church2017agricultural}. However, this advantage is short-lived, as CNN+LSTM fails to maintain its lead in other years. Our fusion-based architecture—although generally more robust—may have underfit certain temporally dominant signals present in that season. This discrepancy highlights a key trade-off between short-term temporal modeling and long-range feature extraction. Additionally, MODIS's lower spatial resolution (250m–1km) increases spectral mixing, which can obscure crop boundaries, reducing the effectiveness of AOL-CNN’s edge detection mechanism. This effect may have been exacerbated in 2013 if atmospheric disturbances (e.g., cloud cover, sensor noise) further degraded spectral quality. Another potential factor is concept drift, where environmental changes in 2013 (e.g., unusual weather patterns affecting crop reflectance) may have led to deviations in spectral distributions, affecting ViT's generalization ability. Despite this isolated underperformance, our model maintains higher temporal stability across multiple years, indicating that AOL-CNN+ViT offers stronger generalization capabilities compared to fixed-feature CNNs, LSTMs, and GP-based approaches. 

Other approaches, such as GNN+RNN demonstrates moderate error rates with occasional competitiveness but lacks temporal consistency, likely due to limited spatial granularity. ConvLSTM exhibits unstable performance across all metrics, struggling to balance spatial and temporal learning, particularly under spectral noise. CNN+GP and LSTM+GP suffer from poor average performance, reflecting the limitations of Gaussian Processes in large-scale, temporally dynamic datasets. DeepYield, while occasionally comparable to better models, lacks robustness and performs inconsistently across years. 3DCNN+ConvLSTM fails to learn meaningful spatiotemporal dependencies, particularly in 2013, and is consistently outperformed. ConvLSTM+ViT, although promising in concept, shows limited synergy in practice, potentially due to suboptimal module interaction or training instability. These observations reinforce that DFYP’s robust design—fusing adaptive edge-aware CNN with ViT-based global modeling—is essential to achieving both accuracy and temporal stability in yield prediction from noisy, low-resolution imagery.

\noindent \textbf{Analysis on Sentinel-2.} Figure~\ref{fig: comparision with sota methods.}(d), (e), (f)  illustrates the yield prediction performance across four major crops—corn, cotton, soybean, and winter wheat—on the Sentinel-2 dataset. Our proposed DFYP model achieves the best performance in terms of RMSE and MAE across all crops, demonstrating its superiority in minimizing absolute prediction errors. For the $R^2$ metric, DFYP consistently outperforms all baselines on corn, soybean, and winter wheat, while achieving the second-best performance on cotton, marginally behind MMST-ViT. This overall trend highlights DFYP's ability to maintain high predictive accuracy and strong variance explanation across diverse crop types.  

We attribute this superiority to three key factors. First, the higher spatial resolution (10m) of Sentinel-2 reduces mixed-pixel effects, enabling AOL-CNN to extract more precise crop boundaries and structural information. In contrast, models trained on lower-resolution inputs suffer from increased spectral mixing—where a single pixel contains signals from multiple land covers—which degrades their ability to delineate precise crop boundaries and spatial structures. Second, the richer spectral information provided by Sentinel-2’s 13-band multispectral data enhances the effectiveness of the SE module’s spectral reweighting, allowing it to prioritize the most relevant spectral channels for yield prediction. This is particularly advantageous compared to CNN-LSTM and GP-based models, which process spectral inputs statically without adaptive reweighting. Third, ViT's ability to model long-range dependencies is is especially effective with high-resolution imagery, capturing large-scale spatial patterns and seasonal variation, which local receptive fields in CNNs or sequential constraints in LSTMs fail to exploit.
Compared to this hybrid and adaptive design, baseline models exhibit various limitations. ConvLSTM+ViT, though enhanced by ViT, lacks edge-specific spatial encoding, weakening its boundary representation. 3DCNN+ConvLSTM and DeepYield rely on rigid feature extractors, leading to overfitting or poor generalization. CNN+GP, LSTM+GP, and CNN+LSTM show moderate performance due to restricted modeling capacity. ConvLSTM ranks lowest across all metrics, underscoring its inability to handle complex spatial-spectral dynamics.

\noindent\textbf{Summary.} The experimental results on both MODIS and Sentinel-2 datasets demonstrate the robustness and adaptability of the proposed DFYP framework across distinct remote sensing conditions. On the MODIS dataset, which features coarse spatial resolution and frequent spectral variability, DFYP achieves consistent performance over multiple years, effectively mitigating temporal drift and sensor noise. On the Sentinel-2 dataset, characterized by high spatial and spectral resolution, DFYP delivers state-of-the-art results across diverse crop types, highlighting its ability to generalize across heterogeneous data distributions. These findings underscore the effectiveness of our dual-branch architecture and adaptive operator design in capturing both local and global patterns under varying spatial, spectral, and temporal conditions. Together, they validate DFYP's potential as a general-purpose solution for robust yield prediction in dynamic agricultural environments.
\subsection{Effectiveness of the Adaptive Operator Learning}
\begin{figure}[h]
  \centering
    \includegraphics[width=1\linewidth]{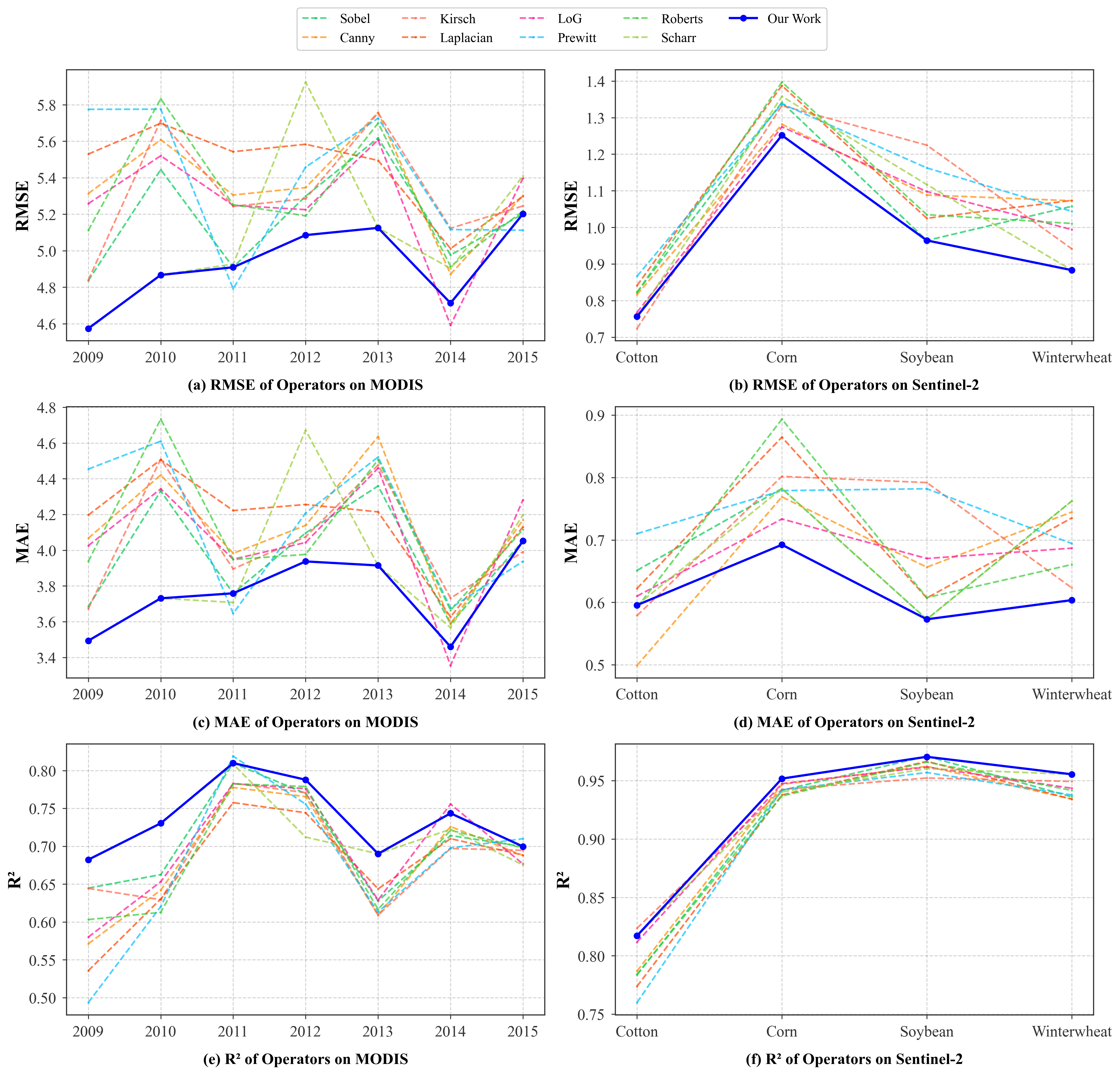}
    \caption{Performance of operators on MODIS and Sentinel-2. Comparison between our adaptive operator library and eight fixed operators}
  \label{fig: Performance of operators}
\end{figure}

Prior to finalizing the design of AOL, we conducted a systematic investigation into how different edge enhancement strategies affect model performance. We conducted a series of experiments by individually integrating eight classical edge operators—Sobel\cite{sobel19683x3}, Canny\cite{canny1986computational}, Kirsch\cite{kirsch1971computer}, Laplacian\cite{marr1980theory}, Log~\cite{marr1980theory}, Prewitt~\cite{prewitt1970object}, Roberts~\cite{do1961machine}, and Scharr~\cite{scharr2004optimal}—into our dual-branch fusion framework. In each case, the selected operator was fixed throughout the training and testing process, while the rest of the architecture remained unchanged. The Results are shown in Figure~\ref{fig: Performance of operators}.

On the MODIS dataset (Figures~\ref{fig: Performance of operators} (a), (c) and (e)), fixed operators exhibited noticeable performance fluctuations across years. RMSE and MAE values varied widely for each operator, indicating their sensitivity to temporal drift.No single operator achieves the best performance across all years and metrics. In contrast, AOL consistently maintains lower RMSE, lower MAE, and higher $R^2$ values, demonstrating its ability to dynamically select the most effective operator based on data characteristics. The stability of AOL across multiple years suggests that an adaptive selection mechanism is necessary to handle the spectral, structural, and environmental variations present in remote sensing imagery. 

On the Sentinel-2 dataset (Figures~\ref{fig: Performance of operators} (b), (d), (f)), performance differences among fixed operators were less pronounced, likely due to the higher spatial resolution and more stable edge characteristics. Nevertheless, our work maintains the lowest RMSE and MAE across most of the crop types, and achieves the highest $R^2$ values, especially for corn, soybean and winterwheat. This demonstrates that AOL can also adapt effectively to cross-crop variability.

Among all the results, Sobel and Scharr emerge as the most frequently top-performing operators, suggesting their broader adaptability under varying temporal and environmental conditions. Motivated by this observation, we developed the Adaptive Operator Learning (AOL), which dynamically selects or learns from these base operators. The experimental results confirm that our dynamic selection strategy not only stabilizes model performance but also leads to consistent improvements in prediction accuracy over most of years and crops. This validates the necessity of an adaptive edge representation mechanism in remote sensing-based yield prediction.

\subsection{Network Architecture Analysis}
To validate the rationale behind Dynamic Fusion, we compare it with three alternative CNN and ViT fusion methods: Hierarchy, CoATNet, and Sequence, as discussed in a comprehensive review \cite{yunusa2024exploring}. The Hierarchy method integrates features in a layered manner, CoATNet leverages attention mechanisms, and Sequence applies a sequential fusion approach. This experiment aims to identify the optimal configuration for balancing local and global feature extraction.

\begin{figure}[h]
  \centering
    \includegraphics[width=1\linewidth]{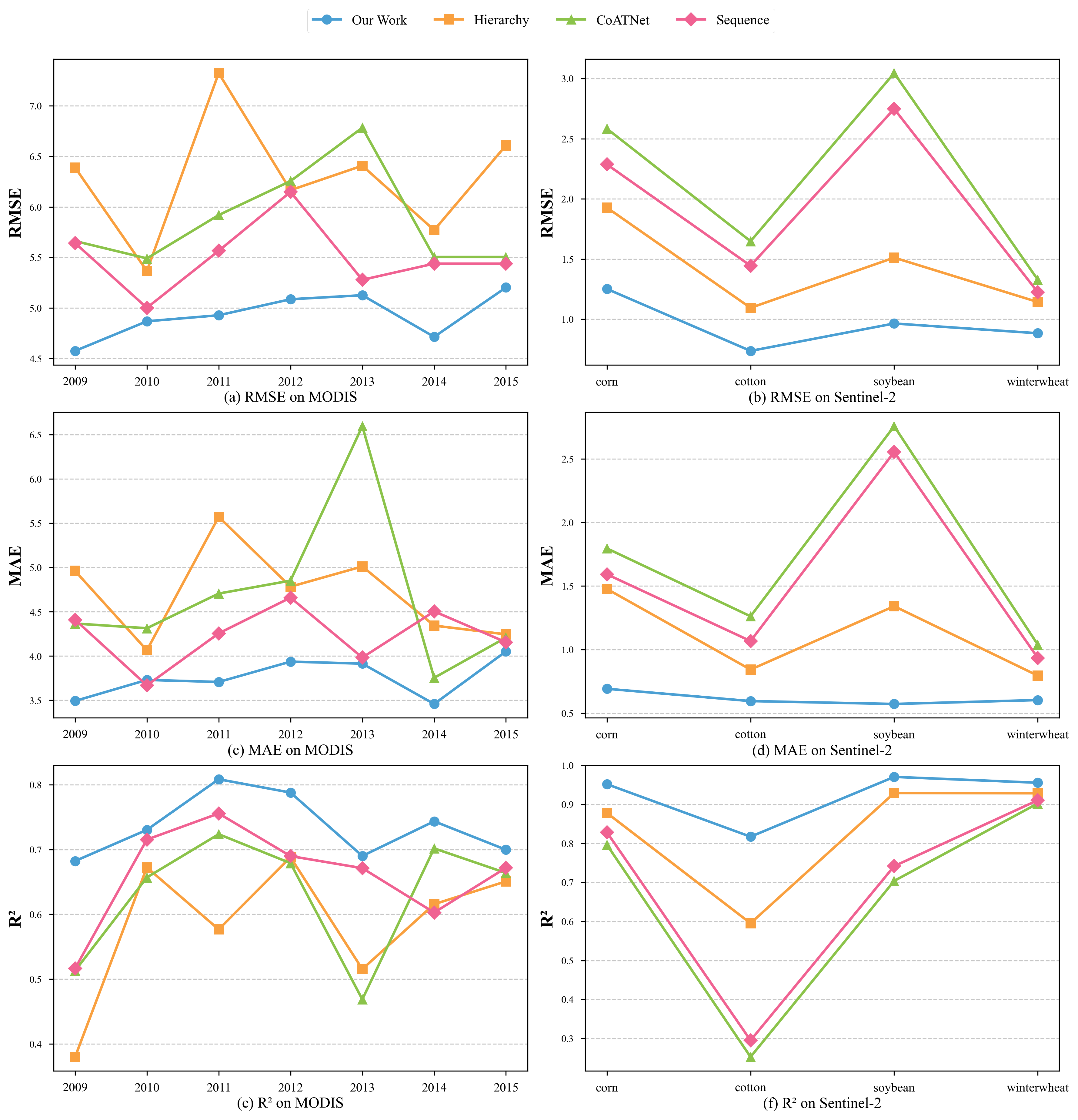}
    \caption{Performance of Different Network Architectures}
  \label{fig:Different Network Architectures}
\end{figure}

The results shown in Figure~\ref{fig:Different Network Architectures} demonstrate that our CNN-ViT fusion approach consistently outperforms alternative fusion strategies for crop yield prediction on both two datasets. Obviously, our method performed very stably on several metrics of the two datasets, which verified its robustness. On MODIS, CoATNet shows large fluctuations, particularly in 2013 and 2014, where both RMSE and MAE spike, and $R^2$ significantly drops. Hierarchy and Sequence methods exhibit moderate performance, but still show more volatility across years compared to ours. On the Sentinel-2 dataset, our method again outperforms all baselines across all crop types. It achieves the lowest RMSE and MAE, and the highest $R^2$ for each crop, including corn, cotton, soybean, and winter wheat. The performance gap is especially noticeable for soybean, where the competing models (particularly CoATNet and Sequence) experience significant degradation in both RMSE and $R^2$. Furthermore, while other models show sharp performance drops when moving across crop types, our method demonstrates minimal fluctuation, indicating strong cross-crop generalization.

To gain deeper insights into these observations, we analyze the underlying mechanisms behind DFYP’s robust performance. Instead of relying on simple concatenation or fixed-weight averaging, DFYP employs a learnable fusion strategy that dynamically adjusts the contributions of CNN and ViT features via trainable coefficients. This enables the model to effectively integrate local textures captured by CNN and global contextual patterns extracted by ViT. While CNN tends to struggle with long-range dependencies due to its limited receptive field, and ViT may lack sufficient fine-grained spatial detail, especially in scenarios with sparse or heterogeneous textures, their fusion compensates for each other’s limitations. This advantage is particularly evident on low-resolution imagery such as MODIS, where capturing both spatial precision and contextual scope is critical. These results highlight the value of an adaptive, data-driven fusion mechanism in improving generalization for remote sensing-based yield prediction.

Overall, the dynamic fusion design in DFYP achieves a well-balanced representation across spatial and semantic dimensions, enabling it to outperform competing architectures and generalize effectively across diverse crops, years, and sensor characteristics.

\subsection{Hyperparameter Analysis}
\label{subsec:hyperparameter_analysis}

To ensure the stability and generalizability of our model, we conducted a comprehensive hyperparameter sensitivity analysis focusing on three key parameters: the number of CNN layers, attention heads, and ViT layers. Table~\ref{tab:rmse_modis} illustrates the impact of these parameters on model performance, evaluated using RMSE.

Our analysis revealed dataset-specific optimal configurations. For the MODIS dataset, the optimal configuration consists of 6 CNN layers, 8 attention heads, and 4 ViT layers, while the Sentinel-2 dataset performs best with 4 CNN layers, 6 attention heads, and 6 ViT layers. These findings demonstrate the importance of dataset-specific optimization and the robustness of our model across different configurations.

\begin{table*}[t]
\centering
\caption{Prediction Error (RMSE) under Different Model Parameters on MODIS Dataset}
\resizebox{\textwidth}{!}{%
\begin{tabular}{c|ccccc|ccccc|ccccc}
\toprule
\multirow{2}{*}{Year} &
\multicolumn{5}{c|}{\textbf{CNN Layers}} &
\multicolumn{5}{c|}{\textbf{Attention Heads}} &
\multicolumn{5}{c}{\textbf{ViT Layers}} \\
& 3 & 4 & 5 & 6 & 7 & 6 & 7 & 8 & 9 & 10 & 2 & 3 & 4 & 5 & 6 \\
\midrule
2009 & 7.2046 & 4.9645 & \textbf{4.5269} & 4.5741 & 5.0481 & 4.7310 & 6.5315 & \textbf{4.5741} & 7.4279 & 4.8407 & 8.4290 & 6.4330 & \textbf{4.5741} & 6.2175 & 9.0705 \\
2010 & 7.1061 & 5.7917 & 5.6418 & \textbf{4.8674} & 5.6889 & 5.4410 & 7.9095 & \textbf{4.8674} & 7.5233 & 5.3096 & 8.6463 & 7.7686 & \textbf{4.8674} & 8.2297 & 8.7567 \\
2011 & 7.6003 & 4.9587 & \textbf{4.8870} & 4.9272 & 5.5756 & 4.9435 & 6.6165 & \textbf{4.9272} & 6.3086 & 5.0740 & 7.7478 & 5.6117 & \textbf{4.9272} & 6.4087 & 7.0807 \\
2012 & 6.0393 & 5.6719 & 5.3731 & 5.0856 & \textbf{5.0538} & \textbf{5.0072} & 7.2861 & 5.0856 & 6.2855 & 5.1956 & 6.8399 & 6.6957 & \textbf{5.0856} & 6.2732 & 6.5589 \\
2013 & 9.9019 & 5.4120 & \textbf{5.0768} & 5.1253 & 5.8255 & 5.2381 & 10.4084 & \textbf{5.1253} & 7.2302 & 5.4257 & 8.8453 & 8.6216 & \textbf{5.1253} & 7.4302 & 10.5766 \\
2014 & 5.9164 & 5.1649 & 4.9827 & \textbf{4.7132} & 4.8971 & 4.9909 & 5.9166 & \textbf{4.7132} & 5.8808 & 4.8708 & 7.0814 & 5.6472 & \textbf{4.7132} & 5.8529 & 6.1671 \\
2015 & 7.4901 & \textbf{5.1520} & 5.4335 & 5.2020 & 5.6517 & 5.2968 & 7.8208 & \textbf{5.2020} & 8.2081 & 5.2886 & 8.9127 & 7.5832 & \textbf{5.2020} & 9.9667 & 10.8091 \\
\bottomrule
\end{tabular}%
}
\label{tab:rmse_modis}
\end{table*}

 \subsection{Ablation Study}

 \begin{figure*}[t]
  \centering
  \includegraphics[width=\textwidth]{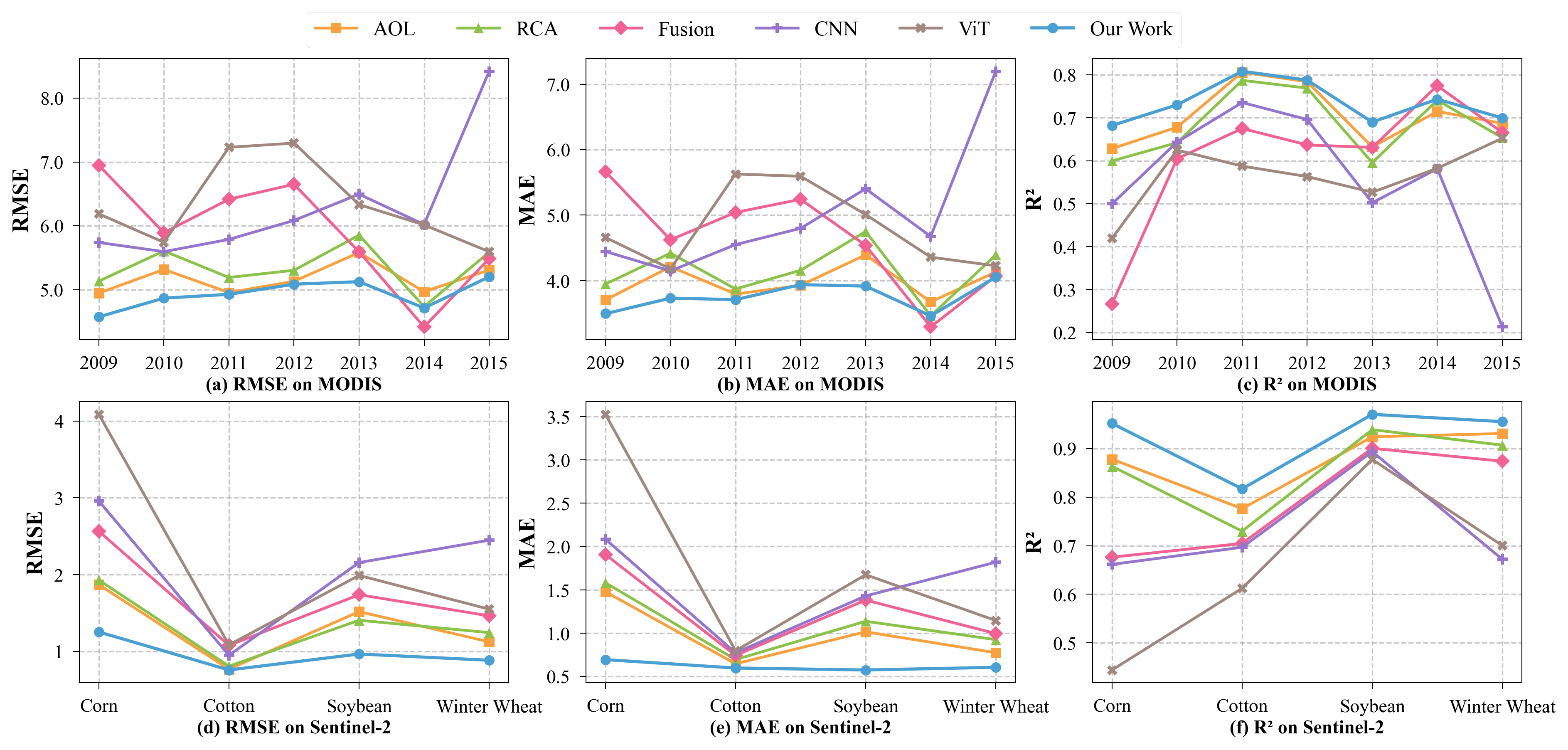}
  \caption{Ablation study results of different components on the MODIS and Sentinel-2 dataset.}
  \label{fig:ablation_sentinel2}
\end{figure*}

To evaluate the contribution of each component, We conducted ablation experiments to investigate the individual and joint contributions of CNN, ViT, the Resolution-aware Channel Attention (RCA), and the Adaptive Operator Learning (AOL) module: (1) CNN-only (CNN), (2) ViT-only (ViT), (3) CNN+ViT Fusion (Fusion), (4) Fusion with AOL (AOL), (5) Fusion with RCA, and (6) Full model (Our work). 

\begin{table}[t]
\centering
\caption{Ablation Study Results on MODIS Data}
\label{tab:ablation study}
\begin{tabular}{ccccccc}
\toprule
CNN & ViT & RCA & AOL & RMSE$\downarrow$ & MAE$\downarrow$ & $R^2$$\uparrow$ \\
\midrule
\checkmark &  &  &  & 6.3061 & 5.0689 & 0.5531 \\
 & \checkmark &  &  & 6.3424 & 4.8069 & 0.5651 \\
\checkmark & \checkmark &  &  & 5.9145 & 4.6377 & 0.6079 \\
\checkmark & \checkmark & \checkmark &  & 5.3443 & 4.1412 & 0.6841 \\
\checkmark & \checkmark &  & \checkmark & 5.1726 & 3.9756 & 0.7045 \\
\checkmark & \checkmark & \checkmark & \checkmark & \textbf{4.9278} & \textbf{3.7563} & \textbf{0.7345} \\
\bottomrule
\end{tabular}
\end{table}

Table~\ref{tab:ablation study} presents abation results on MODIS, while Figure~\ref{fig:ablation_sentinel2} visualizes performance across different crop types in Sentinel-2 and results across different years on MODIS. 

From Table~\ref{tab:ablation study}, we can note that our final model DFYP achieves the best performance with an RMSE OF 4.9278, MAE of 3.7563, and $R^2$ of 0.7345. Compared to the baseline CNN model, DFYP improves the RMSE by 21.86\%, reduces MAE by 25.90\%, and increases $R^2$ by 32.80\%. Similar trends are observed when comparing against other ablation variants. Notably, the inclusion of both AOL and ViT with CNN substantially boosts performance, while the addition of RCA further enhances spatial feature representation.

\noindent \textbf{Discussion on MODIS.} In Figure~\ref{fig:ablation_sentinel2} (a), (b) and (c), our work consistently achieves the lowest RMSE and MAE, and highest $R^2$ in most years, with minimal year-to-year fluctation. It is worth noting that both CNN and ViT individually exhibit considerable performance fluctuations across years, indicating limited temporal robustness. Although the fusion of CNN and ViT improves overall accuracy, it still suffers from noticeable instability—for instance, it performs the worst among all methods in 2009, yet achieves the best result in 2014. This suggests that simple backbone fusion does not fully address the temporal drift in remote sensing data. The addition of either the RCA or AOL module mitigates these fluctuations to some extent. Specifically, integrating AOL yields more significant improvements in both accuracy and stability compared to RCA, emphasizing the greater importance of adaptive edge features over spectral-channel attention in yield prediction. Ultimately, the full DFYP model—incorporating both RCA and AOL—achieves the best performance and the most consistent results across years, demonstrating its effectiveness in handling temporal variability and concept drift in remote sensing-based yield estimation.

\noindent \textbf{Discussion on Sentinel-2.} On the Sentinel-2 dataset (shown in Fig.~\ref{fig:ablation_sentinel2} (d), (e) and (f)), our fusion-based model demonstrates superior performance and stability across all four crop types. Notably, the individual CNN and ViT models exhibit substantial fluctuations across the three evaluation metrics (RMSE, MAE, and $R^2$), with ViT showing particularly high variance. By fusing CNN and ViT, both performance and stability are improved to some extent, suggesting that combining local and global features is beneficial. Further enhancements are observed when either the AOL or RCA module is integrated into the fusion model, with both additions leading to noticeable improvements in accuracy and consistency.

Interestingly, unlike on the MODIS dataset, the performance gap between the AOL- and RCA-enhanced models is relatively small on Sentinel-2, especially in terms of RMSE. However, a closer look at the MAE and $R^2$ metrics reveals that AOL still provides more pronounced benefits than RCA, reinforcing the value of edge-aware adaptive feature extraction. When both AOL and RCA are incorporated—yielding our full DFYP model—the resulting performance and stability reach their highest levels, as evidenced by the consistently superior blue lines across all metrics.

\noindent \textbf{Summary.} These observations confirms that each component—AOL, RCA, and ViT—enhances model performance, with their combination yielding the best results. ViT improves long-range feature extraction in low-resolution data (MODIS), AOL enhances edge features in high-resolution data (Sentinel-2), and RCA stabilizes spectral feature selection, particularly benefiting MODIS. These results highlight the necessity of a dynamically fused framework for robust crop yield prediction across diverse remote sensing conditions. Supplementary material is visible for detailed analysis of each component on two datasets. We summarize the results of the ablation experiments on the two datasets in Table~\ref{tab:AOL_SEB_ViT}.

\begin{table*}[t]
\footnotesize 
\centering
\caption{Comparative Impact of AOL, RCA, and ViT Across MODIS and Sentinel-2 Datasets.}
\label{tab:AOL_SEB_ViT}
\setlength{\tabcolsep}{3pt} 
\begin{tabular}{p{2.5cm}p{7.25cm}p{7.25cm}} 
\toprule
\textbf{Component} & \textbf{MODIS (Low Resolution, Multi-Year)} & \textbf{Sentinel-2 (High Resolution, Diverse Crops)} \\
\midrule
\textbf{AOL} & 
Enhances edge features, compensating for low spatial resolution.  
Reduces temporal inconsistencies across years.  
Fixed operators struggle with spectral variations. &  
Most crucial for high-resolution imagery.  
Improves CNN’s ability to capture fine-grained crop boundaries.  
Key for preserving local structure. \\
\midrule
\textbf{RCA} & 
Essential for handling limited spectral bands.  
Reweights spectral channels to reduce spectral noise.  
Improves multi-year prediction stability. &  
Still beneficial, but less impactful due to richer spectral data.  
Aids spectral selection but plays a secondary role in feature extraction. \\
\midrule
\textbf{ViT} &  
Most effective in low-resolution settings.  
Captures long-range dependencies, compensating for resolution constraints.  
Improves feature aggregation for multi-year predictions. &  
Beneficial but less dominant, as AOL-Net extracts local features well.  
Supports large-scale pattern recognition but is less critical in high-res imagery. \\
\midrule
\textbf{Conclusion} &  
ViT plays the most crucial role in low-resolution data, compensating for spatial limitations.  
AOL stabilizes edges, while RCA mitigates spectral inconsistencies. &  
AOL is the primary driver of improvement in high-resolution data.  
RCA helps but is less impactful, while ViT provides complementary spatial learning. \\
\bottomrule
\end{tabular}
\vspace{-3mm} 
\end{table*}
\subsection{More Results: Spatial Visualization}
\label{sec: Spatial Visualization}

\begin{figure*}[h]
  \centering
    \includegraphics[width=1\linewidth]{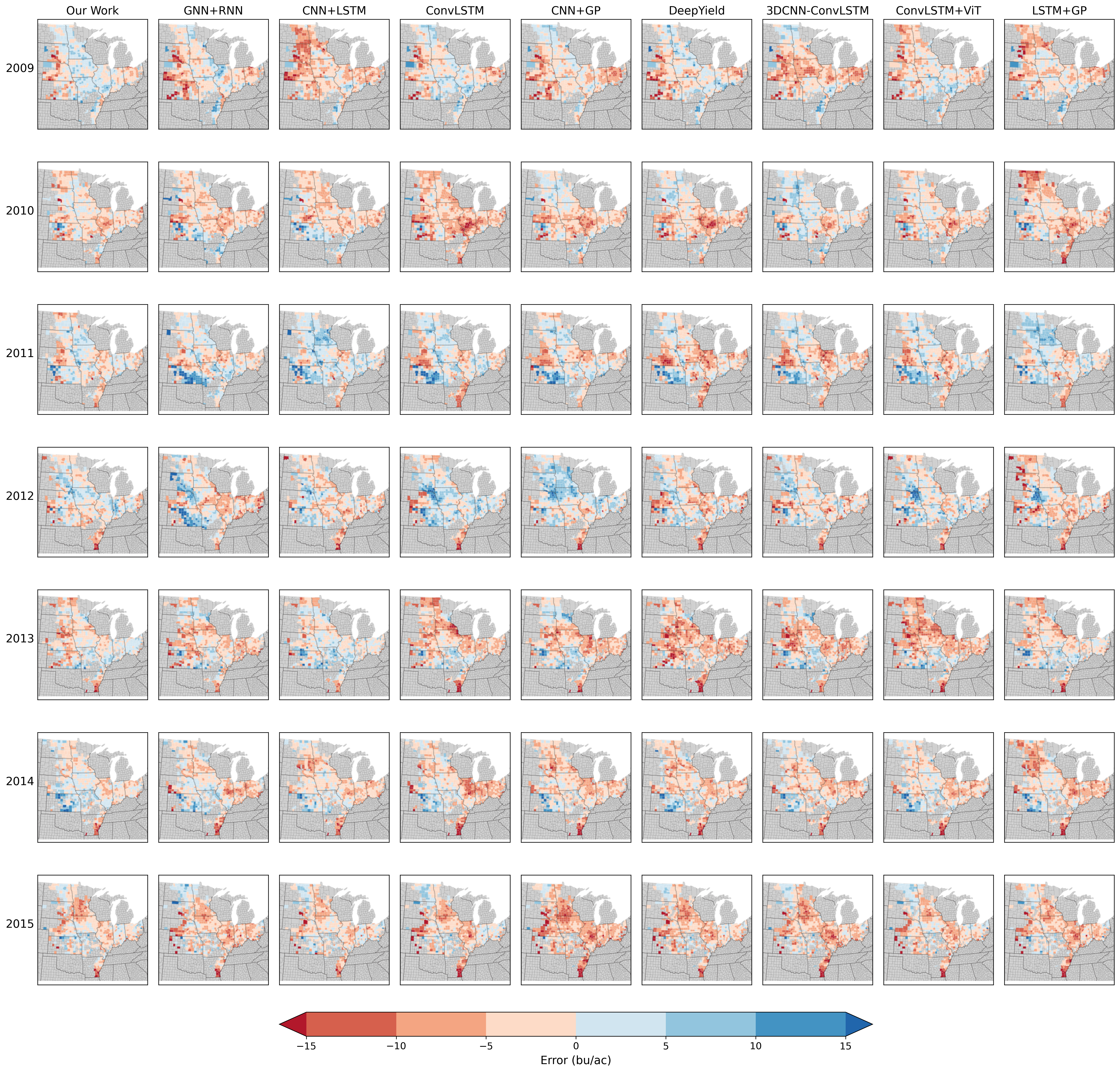}
  \caption{Spatiotemporal Error Maps for Comparative Model Analysis}
  \label{fig:combined_map}
\end{figure*}

To complement the quantitative metrics aforementioned, we utilize spatiotemporal error maps to demonstrate a more intuitive comparison between our method and existing state-of-the-art approaches, which are shown in Figure~\ref{fig:combined_map}. Each map visualizes county-level yield prediction errors over time, where colors range from deep red (severe underestimation, -15 bu/ac) to deep blue (severe overestimation, +15 bu/ac). Lighter tones near the center indicate predictions closer to ground truth.

Compared to other methods, DFYP consistently exhibits a more neutral and spatially balanced error distribution across all years, with significantly fewer regions showing extreme under- or overestimation. In contrast, baseline models such as CNN+LSTM, ConvLSTM, and DeepYield display larger spatial error clusters—particularly in regions with complex terrain or heterogeneous crop conditions—indicating limited generalization capacity under diverse environmental settings.

This visualization explicitly highlights the advantage of the proposed framework in both accuracy and robustness. The reduced spatial error variance suggests that the combination of adaptive edge-aware feature extraction (via AOL-CNN) and global context modeling (via ViT) enables DFYP to better handle regional variability and reduce systematic bias. Moreover, our method maintains stable performance even in climatically volatile years (e.g., 2012–2013), demonstrating its resilience to temporal disturbances. This resilience under climatic extremes suggests that DFYP is not only accurate under ideal conditions, but also robust and generalizable in real-world, high-variance agricultural environments. 

\section{Conclusion}
We propose DFYP, a dynamically fused crop yield prediction framework integrates adaptive feature extraction and multimodal fusion to enhance robustness across diverse remote sensing datasets. Our approach introduces three key innovations: a dual-branch architecture that optimally balances local and global feature extraction through learnable weight coefficients, an Adaptive Operator Learning (AOL) that selects the most suitable edge detection operator (Sobel, Scharr, or a learnable kernel) based on historical performance to improve spatial feature extraction, and a Resolution-Aware Channel attention (RCA) module that applies max pooling after histogram processing for low-resolution data (e.g., MODIS) to enhance key spectral features while using average pooling for high-resolution data (e.g., Sentinel-2) to preserve fine spatial details. Extensive experiments on MODIS and Sentinel-2 datasets demonstrate that DFYP consistently outperforms state-of-the-art models across multiple evaluation metrics. The results validate DFYP'S adaptability to varying spatial resolutions, environmental conditions, and crop types. Additionally,  DFYP demonstrates robustness under real-world constraints, making it a practical and scalable solution for modern agricultural monitoring and decision support.

In future work, we plan to incorporate uncertainty quantification to enhance predictive reliability and explore more efficient edge detection strategies tailored for high-resolution remote sensing data.

\section*{Acknowledgments}
This paper has received significant assistance from Dr. Jing Zhang of the Australian National University. At the same time, we would like to express our gratitude to the editor and reviewers for their valuable comments on this article.
{\appendices
\section{More Dataset Description}
\label{sec:dataset description}
In the experiment, we employ two datasets, which together contain four crops at different resolutions, to evaluate the performance of the model. The details are provided below.

\noindent\textbf{MODIS data:}
The Moderate Resolution Imaging Spectroradiometer (MODIS) is a key instrument of the Earth Observing System (EOS) onboard NASA’s Terra satellite, launched in 1999, and Aqua satellite, launched in 2002. In this study, we utilize MODIS-derived data, which includes three components: surface reflectance, surface temperature, and soil cover type.

\begin{figure*}[h]
  \centering
  \subfloat[Soybean production by statistical provinces in 2020\label{fig:soybean_2020}]{
    \includegraphics[width=0.48\linewidth]{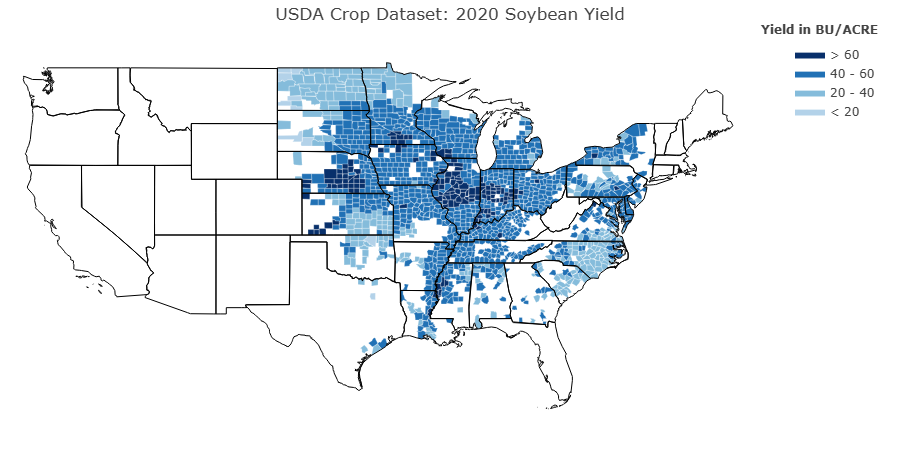}
  }
  \hfill
  \subfloat[Geographical distribution of Sentinel-2 image collections\label{fig:sentinel2_distribution}]{
    \includegraphics[width=0.48\linewidth]{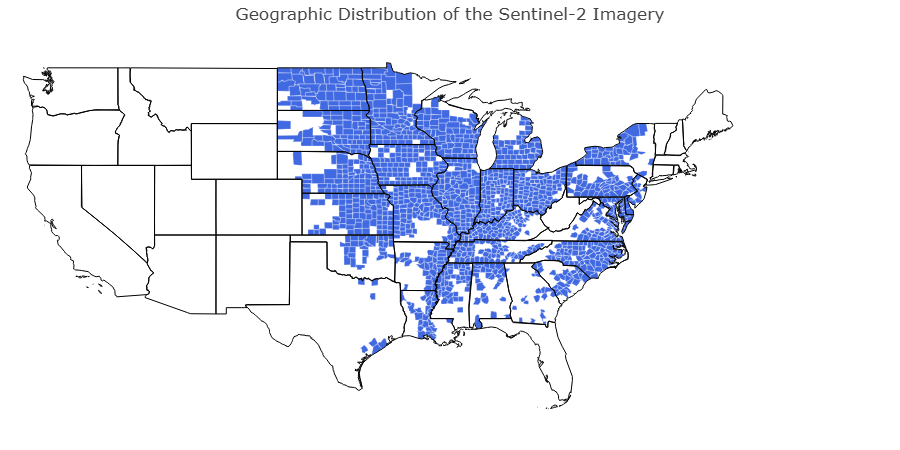}
  }
  \caption{Visualization of the Sentinel-2 dataset.}
  \label{fig:yield_prediction}
\end{figure*}

\emph{1) surface reflectance} The MOD09A1 product, derived from the MODIS instrument onboard the Terra satellite, provides surface reflectance data for bands 1–7, which are similar to those of the Thematic Mapper (TM) sensor but feature narrower bandwidths. This characteristic reduces atmospheric effects and enhances data quality. Various indices can be extracted from reflectance data, such as the water body index, vegetation index, brightness index, and other classification indices, which are widely used in classification methodologies. The reflectance spectra of vegetation, soil, and water bodies exhibit significant differences in the visible and near-infrared (NIR) bands. For instance, in the \textbf{RED band} (0.620--0.670~$\mu$m), vegetation and soil show lower reflectance compared to the NIR band (0.700--1.100~$\mu$m), and the RED band is sensitive to vegetation cover and growth. In the \textbf{BLUE band} (0.459--0.479~$\mu$m), differences between soil and vegetation are pronounced. The \textbf{NIR band} (0.841--0.876~$\mu$m) is characterized by high reflectance in vegetation, making it a strong indicator for vegetation-related analysis. For water bodies, reflectance is primarily observed in the green and BLUE bands, while in the NIR and MIR regions, water bodies exhibit strong absorption of light waves. In contrast, soil either absorbs or reflects incident light depending on its composition and condition.

\emph{2) surface temperature} Surface temperature is influenced by various factors, including sunlight duration, light intensity, local air temperature, and atmospheric humidity. Vegetation cover also plays a significant role, with surface temperatures typically higher in areas lacking vegetation compared to those with abundant vegetation. The MOD11 product provides surface temperature data with varying temporal and spatial resolutions; in this study, we utilize the MOD11A2 product.

\emph{3) soil cover type} The MODIS land cover type product MCD12Q1 provides global land cover data at a spatial resolution of 500 meters. It includes classifications from five different systems, 12 scientific datasets, and a three-layer legend based on the Food and Agriculture Organization’s Land Cover Classification System (LCCS). Additionally, it contains a quality control layer and a binary land-water mask, supporting the creation of land cover maps with six different legends.

\noindent\textbf{Sentinel-2 data:}  
We utilize the Sentinel-2 dataset introduced by Lin et al.~\cite{lin2024open}, which provides high-resolution satellite imagery for monitoring crop growth across the continental United States. This dataset is derived from the Sentinel-2 mission, consisting of twin satellites positioned 180\textdegree{} apart in the same orbit, theoretically enabling a 5-day global revisit cycle. However, due to data availability constraints and cloud coverage considerations, the dataset was constructed with a 14-day sampling interval.  

As described in~\cite{lin2024open}, the dataset was collected using the Sentinel Hub Processing API, accessing Sentinel-2 Level-1C (L1C) imagery with a maximum cloud coverage threshold of 20\%. Each image in the dataset has a spatial dimension of 224$\times$224 pixels, covering a 9$\times$9 km ground area. Instead of standard RGB bands, three specific spectral bands were selected: B02 (blue), B08 (near-infrared/NIR), and B11 (short-wave infrared/SWIR), which are particularly valuable for agricultural monitoring. While the native resolutions of these bands vary (10m for B02 and B08, 20m for B11), all images were resampled to a uniform spatial resolution for consistency.  

The dataset spans six years (2017-2022) and covers 2,291 counties across the continental United States, providing extensive spatial and temporal coverage for analyzing crop development patterns. Figure~\ref{fig:yield_prediction} presents a spatial overlay of Sentinel-2 image coverage and county-level soybean yield data from 2020, illustrating the alignment between satellite observations and agricultural production.  
\begin{figure}[h]
  \centering
   \includegraphics[width=1\linewidth]{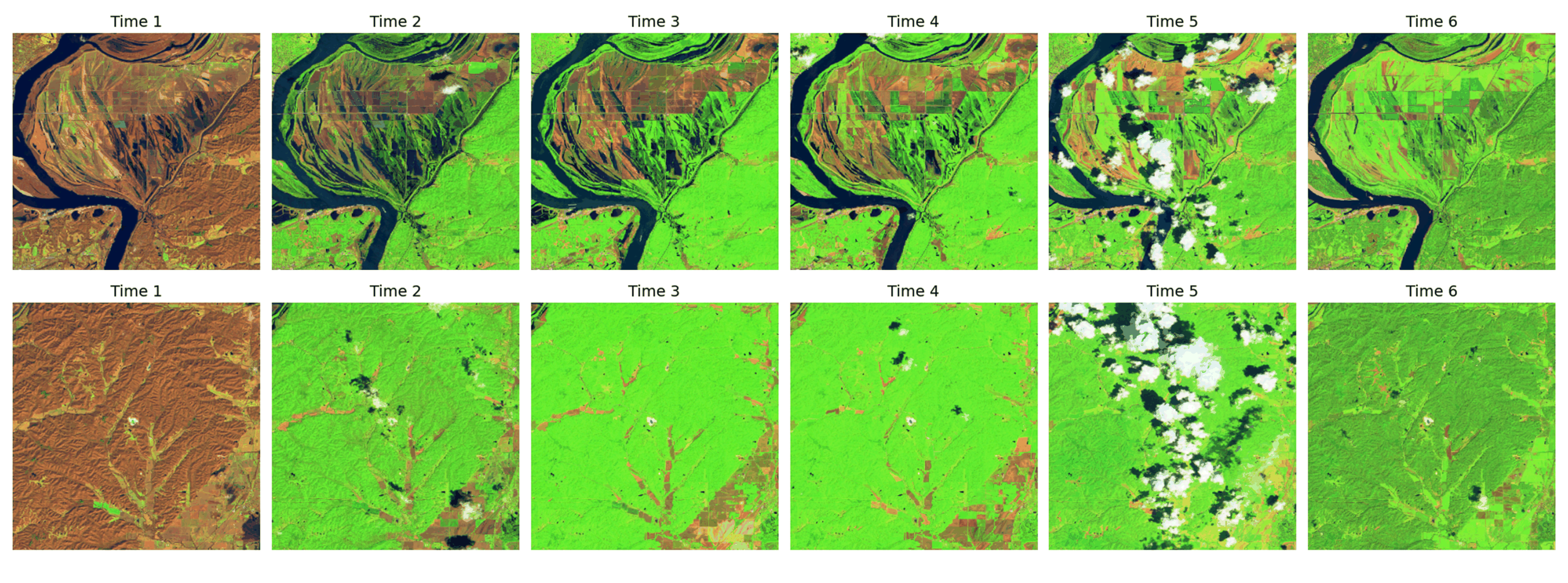}
   \caption{Example image of Sentinel-2 data with different time.}
   \label{fig:Example image of Sentinel-2 data}
\end{figure}
\begin{figure}[h]
  \centering
   \includegraphics[width=1\linewidth]{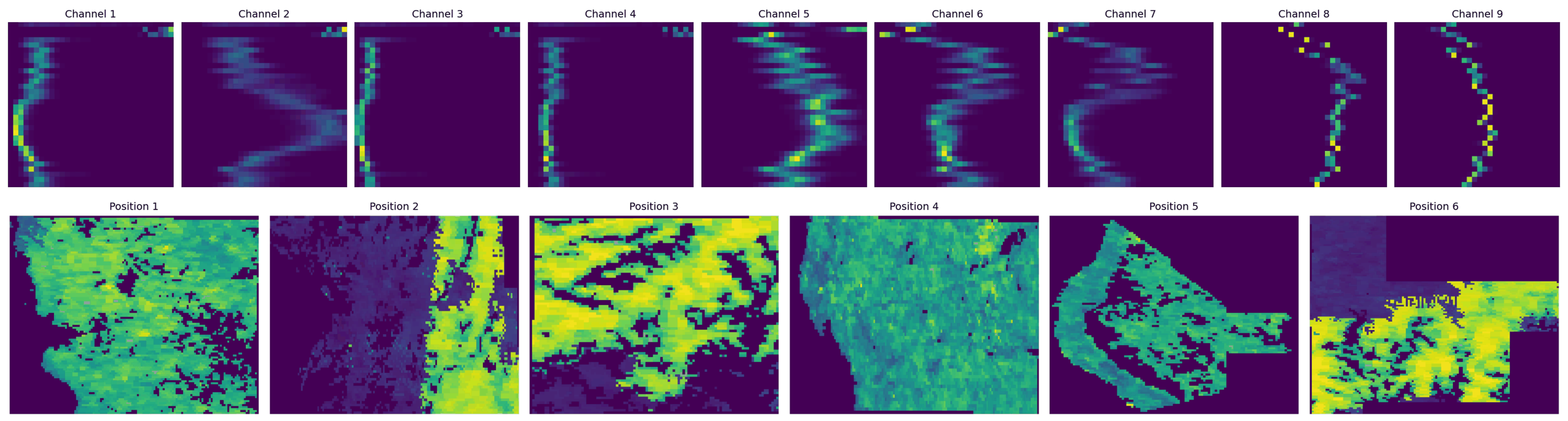}
   \caption{Histogrammed image of MODIS data. The upper row shows the visualizaition of 9 channels histogrammed images. The bottom row indicates 6 position images of MODIS data.}
   \label{fig:all channels of histogrammed images}
\end{figure}

\section{Data Processing}
\label{sec:data processing}
To enhance the accuracy of crop yield prediction, we applied specific preprocessing methods to datasets with varying resolutions to extract high-quality information. For the Sentinel-2 dataset from Lin et al.~\cite{lin2024open}, we utilized the data directly as it was already appropriately processed with standardized image dimensions, cloud filtering, and spectral band selection as described in the previous section. While no additional preprocessing was required for this dataset, we visualized sample images to verify data quality and characteristics, as shown in Figure~\ref{fig:Example image of Sentinel-2 data}.

For MODIS data, which includes surface reflectance (7 bands), surface temperature (2 bands), and land cover type (1 band) from 2009 to 2015, preprocessing was necessary due to the varying temporal resolutions: surface reflectance and temperature are available every 8 days, while land cover data are annual. The preprocessing steps are as follows:

\emph{1) Annual Classification and Spectral Extraction} Firstly, we classify the MODIS data by year, extract the required spectral data for each year to prepare for further processing. Let \emph{B} represent the total of stacked channels, \emph{N} indicate the number of channels per acquisition of the spectral product, and \emph{M} be the acquisition interval in days. The specific number of spectral bands per year is calculated as follows:
\begin{equation}
  bands_{1year}= N \times \lfloor \frac{365}{M} \rfloor
  \label{eq:bands}
\end{equation}
where $\lfloor.\rfloor$ indicates the rounding down operator. 

\emph{2) Preprocessing and Reclassification of Land Cover Data} Preprocess the annually extracted land cover data by removing irrelevant information and reclassifying it based on specific criteria. For instance, convert the data into a binary classification by assigning a value of 1 to designated land cover types (e.g., crop areas) and 0 to others. This transformation generates a mask for filtering purposes.

\emph{3) Mask Generation and Application} Based on the reclassified land cover data, generate a specific mask to identify the target areas. Apply this mask to the annually extracted spectral data, effectively retaining relevant information within target areas and filtering out unrelated regions.

\emph{4) Channel Combination} Specifically, the surface reflectance and surface temperature channels for the same area, segmented by year, are stacked together, resulting in a combined spectral channel with 9 channels. This combined spectral data was then further processed using the mask to obtain the final pre-processed spectral data.

\emph{5) Histogram Transformation} In low-to-medium resolution images, local details may be blurry and susceptible to noise. By focusing on the global distribution of data, histograms effectively capture overall features while minimizing noise impact. We transformed spectral images into histograms by calculating pixel frequencies within specified intervals for each spectral band, thereby enhancing feature extraction and robustness. A consistent interval specification was applied across all counties, standardizing the dataset and facilitating seamless input into the neural network. Figure~\ref{fig:all channels of histogrammed images} illustrates the comparison of images before and after histogram transformation. An example of the processed image is shown in Figure~\ref{fig:Example image of Sentinel-2 data}.

In this supplementary material, we provide a comprehensive foundation that supports the main paper's findings, offering deeper insights into the data characteristics and processing methodologies of our DFYP model for crop yield prediction from remote sensing imagery.
}
{
    \small
    \bibliographystyle{IEEEtran}
    \bibliography{main}
}

\end{document}